\documentclass[10pt,journal,compsoc]{IEEEtran}
\ifCLASSOPTIONcompsoc
  \usepackage[nocompress]{cite}
\else
  \usepackage{cite}
\fi
\ifCLASSINFOpdf
   \usepackage[pdftex]{graphicx}
\else
\fi
\usepackage{amsmath}
\usepackage{algorithmic}

\usepackage{multirow}
\usepackage{amsfonts}
\usepackage{caption}

\begin{document}
%
\title{Learning TFIDF Enhanced Joint Embedding for Recipe-Image Cross-Modal Retrieval Service}
%
%
%
%

\author{Zhongwei Xie,
	Ling Liu~\IEEEmembership{Fellow,~IEEE},
	Yanzhao Wu,
        Lin Li,
        and~Luo Zhong
\IEEEcompsocitemizethanks{\IEEEcompsocthanksitem Z Xie, L Liu and Y Wu are with the School of Computer Science, Georgia Institute of Technology, Atlanta, GA 30332.\protect\\
E-mail: zhongweixie@gatech.edu
\IEEEcompsocthanksitem Z Xie, L Li and L Zhong are with the School of Computer Science and Technology, Wuhan University of Technology, China, 430070.}
\thanks{Manuscript received September 24, 2020; revised March 19, 2021.}}

%
%

\markboth{Journal of \LaTeX\ Class Files,~Vol.~14, No.~8, August~2015}%
{Shell \MakeLowercase{\textit{et al.}}: Bare Demo of IEEEtran.cls for Computer Society Journals}
%



\IEEEtitleabstractindextext{%
\begin{abstract}
It is widely acknowledged that learning joint embeddings of recipes with images is challenging due to the diverse composition and deformation of ingredients in cooking procedures. We present a Multi-modal Semantics enhanced Joint Embedding approach (MSJE) for learning a common feature space between the two modalities (text and image), with the ultimate goal of providing high-performance cross-modal retrieval services. Our MSJE approach has three unique features. First, we extract the TFIDF feature from the title, ingredients and cooking instructions of recipes. By determining the significance of word sequences through combining LSTM learned features with their TFIDF features, we encode a recipe into a TFIDF weighted vector for capturing significant key terms and how such key terms are used in the corresponding cooking instructions. Second, we combine the recipe TFIDF feature with the recipe sequence feature extracted through two-stage LSTM networks, which is effective in capturing the unique relationship between a recipe and its associated image(s). Third, we further incorporate TFIDF enhanced category semantics to improve the mapping of image modality and to regulate the similarity loss function during the iterative learning of cross-modal joint embedding. Experiments on the benchmark dataset Recipe1M show the proposed approach outperforms the state-of-the-art approaches.

\end{abstract}

\begin{IEEEkeywords}
Cross-modal retrieval service, Deep learning, Joint embedding learning, TF-IDF, Multi-modal semantics.
\end{IEEEkeywords}}

\maketitle

\IEEEdisplaynontitleabstractindextext

%
\IEEEpeerreviewmaketitle

\IEEEraisesectionheading{\section{Introduction}\label{sec:introduction}}

%
%
%
%
\IEEEPARstart{C}{ross-modal} retrieval is one of the most desired services in big data powered machine learning applications. Many real world datasets have more than one modality, such as Twitter tweets, Instagram messages, yelp restaurant/dish recommendations, Food-101~\cite{Food101}, RecipeIM~\cite{Salvador+CVPR2017_JESR}, and other social platforms (e.g., www.food.com, cookeatshare.com). By training deep neural networks to learn a multi-modal joint embedding model, say a text-image cross-modal embedding model, one can predict the top-k most relevant images given a text document (e.g., finding the top 5 food images best matching a given tweet in Twitter), or the top-k most relevant text documents for a given image (e.g., finding the top 5 tweets best matching a given landslide photo image). Unsupervised learning of multi-modal joint embedding remains challenging~\cite{Lecun-2020AAAi, Yan+CVPR2015} for cross-modal retrieval.

Recently, a few proposals are put forward to address the cross-modal embedding problem~\cite{Food101, Salvador+CVPR2017_JESR, Carvalho+SIGIR2018_AdaMine, JinJinChen+MM2018_AMSR, JinJinChen+MM2017_SAN} using Recipe1M, with over 800K structured cooking recipes (title, list of ingredients, and cooking instructions) and over 1 million associated images~\cite{Salvador+CVPR2017_JESR}. There are several reasons that the problem of learning cross-model embedding for recipes and food images is representative and challenging.
{\em First}, 
Recipe1M is a public and unlabeled real world dataset, well-understood and well-structured, without concerns on user privacy risks. 
{\em Second}, user-generated and shared food images and recipes are prevalent on the Internet and in social media platforms~\cite{Tratter+WWW2017}, and most of them provide ingredients with their quantities, and cooking instructions on how ingredients are prepared and cooked (e.g., steamed, stir-fried, or deep-fried), providing a new source of references for food intake tracking and health monitoring and management, which are not provided in food label or FCT (Federal Capital Territory)~\cite{FCT}. Finally, most existing approaches differ in how they leverage deep neural networks to learn a cross-modal embedding for cooking recipes and food images. However, the reported test accuracy results have been low for both image to recipe retrieval and recipe to image retrieval. The top-1 recall and top-5 recall for image-to-recipe retrieval is 12.5\% and 31\%~\cite{JinJinChen+MM2017_SAN}, 24\% and 51\%~\cite{Salvador+CVPR2017_JESR}, 25.6\% and 53.7\%~\cite{JinJinChen+MM2018_AMSR}, or 39.8\% and 69\%~\cite{Carvalho+SIGIR2018_AdaMine}. 
\begin{center} 
\begin{figure*}
  \centering
  \includegraphics[scale=0.21]{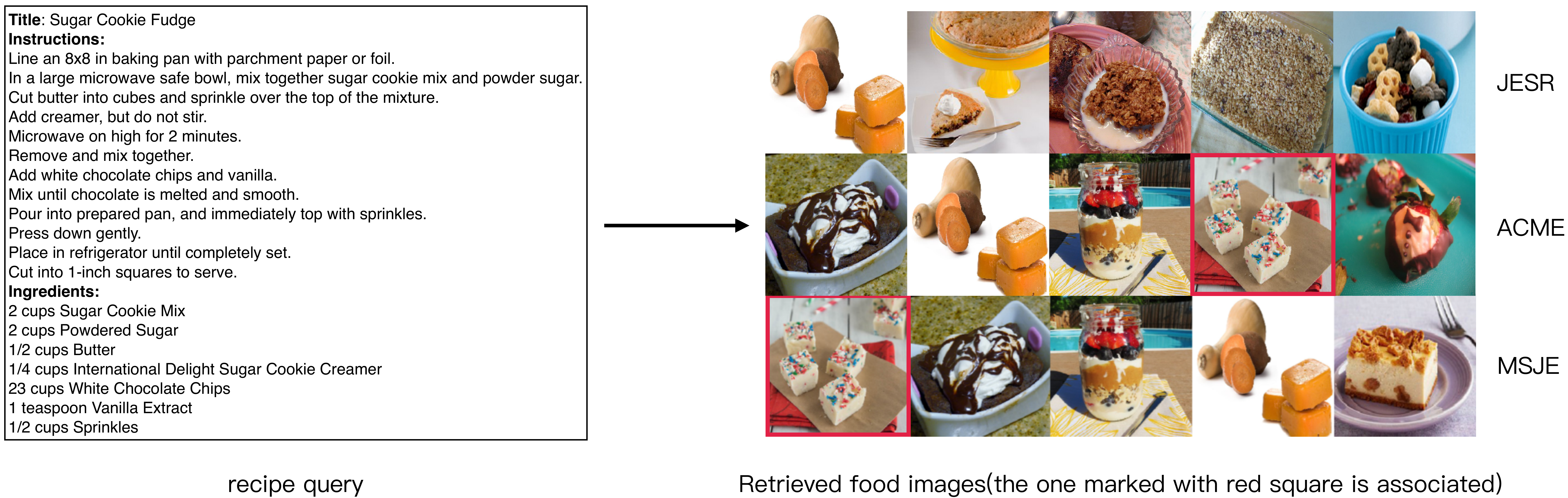}
  \caption{A recipe-to-image retrieval example. }
  \label{retrieval_exp}
\end{figure*}
\end{center} 
Figure~\ref{retrieval_exp} provides an illustrative example, which is a recipe query to find the top-5 best matched food-image instances. The resulting 5 images are sorted by the relevance ranking in descending order. The red frame around an image indicates that it is the original image instance associated with this querying recipe. The first two rows are the top 5 results from two existing approaches and the last row is the top 5 results using our approach reported in the paper, in which the original image instance is included in the top 5 matched images (top 5 hit) and is also ranked as the top-1 matched image (top 1 hit). Unfortunately, most of the existing approaches will fail this recipe-to-image query in either top-5 or top-1 metric, because the original image is not included in the top 5 matched images at all (row 1), or the original image is included in the top 5 matched images but not ranked as the top-1 image (row 2).

In this paper, we present a Multi-modal Semantic enhanced Joint Embedding approach, coined as MSJE, for delivering high-performance cross-modal retrieval services that support image to recipe retrieval task and recipe to image retrieval task. Given that the problem is to learn a multi-modal joint embedding over an unlabeled dataset with each input containing two modalities, such as Recipe1M, by mapping features of different modalities onto the same latent space for similarity-based assessment and recommendation, such as cross-modal retrieval. We argue that 
(1) the quality of the features used from each modality for learning the joint embedding is critical for the effectiveness of learning; and 
(2) the semantic correlations of different modalities can be an effective optimization to regulate the cross-modal similarity loss function used in iterative learning of the cross-modal joint embedding.

This paper makes three main contributions. 
{\em First},  we extract the TFIDF feature from the title, ingredients and cooking instructions of recipes as recipe frequency semantics. By determining the significance of word sequences through combining LSTM learned features with their TFIDF features, we encode a recipe into a TFIDF weighted vector for capturing significant key terms and how such key terms are used in a sequence of sentences in the corresponding cooking instructions. This is in contrast to the existing approaches that treat every keyword and every sentence in a recipe equally when modeling joint text-visual (image) relevance, and overlooking the fact that some descriptions of words and sentences are not visually observable (w.r.t. the corresponding food images).
{\em Second}, in addition to extracting the sequence feature by learning sentence sequences of cooking instructions through two-stage LSTM networks, we incorporate frequency and sequence semantics into the recipe embedding by combining the recipe TFIDF feature and recipe sequence feature. This TFIDF semantics enhanced sequence extraction enables our MSJE approach to effectively learn the unique relationship between the given recipe and its associated food image(s).
{\em Third}, we further incorporate category semantics by combining TFIDF of recipe title with Food-101 classification of food images in both learning the mapping of food image modality and exercising semantic regularization to optimize the similarity loss function during the iterative learning of cross-modal joint embedding. 
We conduct an extensive experimental evaluation and our results show that MSJE outperforms the state-of-the-art approaches with high-quality cross-modal retrieval performance on the benchmark Recipe1M dataset in terms of top-1, top-5, top-10 recall measures for both image2recipe and recipe2image queries. 

\begin{figure*}[!t] 
  \centering
  \includegraphics[scale=0.21]{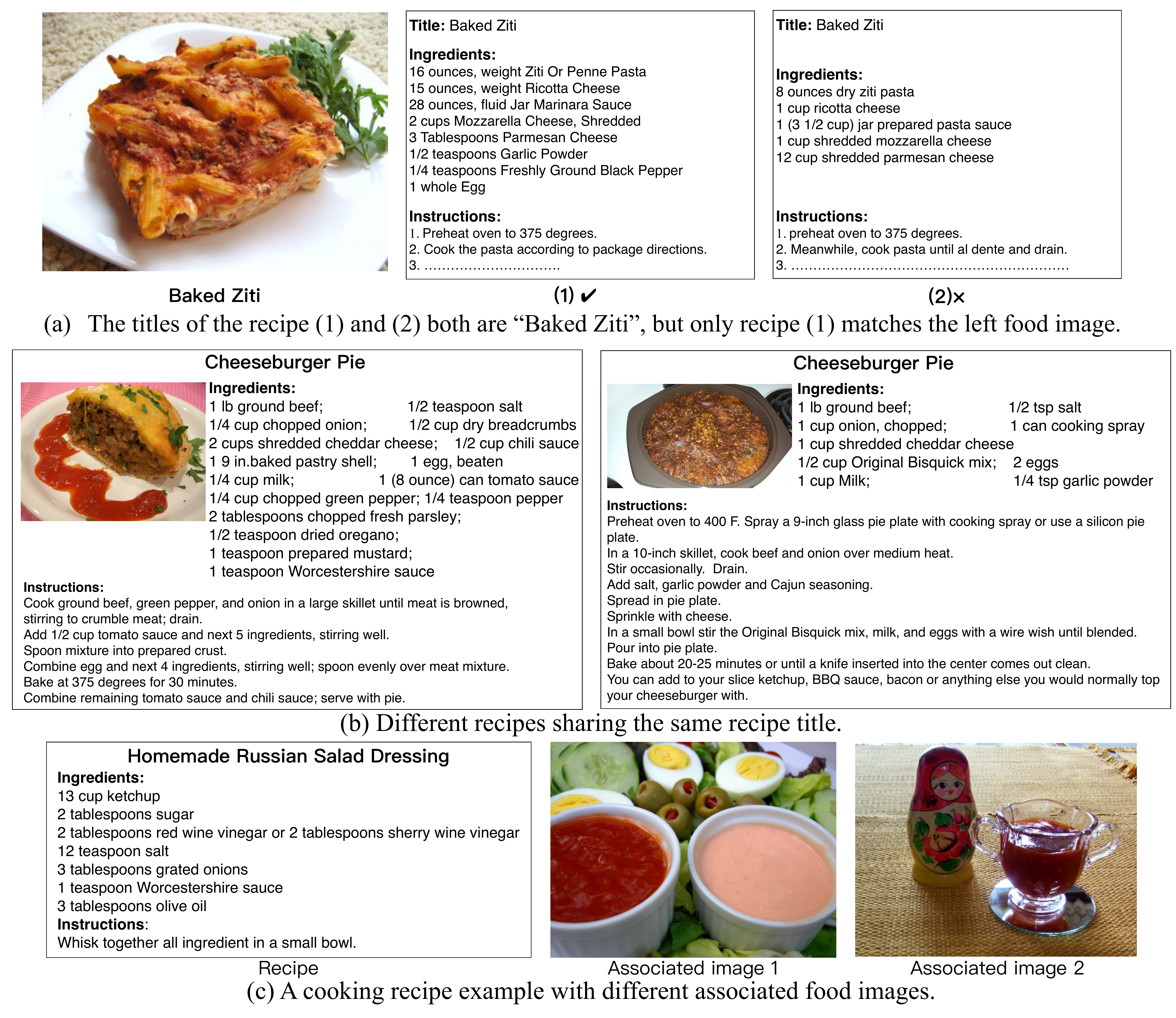}
  \caption{Some recipe-image examples in the Recipe1M dataset .}
  \label{ab-example}
\end{figure*}

The novelty of our proposed MSJE framework centers on a suite of semantic optimizations introduced from multiple dimensions to improve the quality of cross-modal joint embedding learning: (1) We optimize the recipe text embedding using TFIDF features to refine the LSTM key term extraction in both key-term filtering and key-term weights. (2) We optimize the food image embedding using both image category embedding combined with image pixel embedding. (3) We optimize the joint embedding learning by introducing the semantics enhanced loss optimization that combines the batch-hard triplet loss with two semantic optimizations for modality alignment loss and modality independent semantic regularization loss. 
For instance, the titles of the two recipes in Figure~\ref{ab-example}(a) are identical, but only recipe (1) is the correct match to the query image on the left. This can be achieved by incorporating the TFIDF features from title, ingredients, and cooking instruction with (i) the sequence features from LSTM, (ii) the auto encoding of image modality and classification, and (iii) the loss function regulation for learning the joint embedding. To the best of our knowledge, the MSJE approach is unique in employing and combining semantics enhanced feature extraction, feature engineering and feature alignment in single modality embedding learning and cross-modality joint embedding learning.


The remainder of this paper is organized as follows. Section~\ref{relatedwork} outlines the most relevant previous works. Section~\ref{framework} presents the detailed design of MSJE, including how different types of semantics are incorporated in feature engineering of each modality and in iterative learning of cross-modal joint embedding. We report the evaluation results by comparing our approach with the representative state-of-the-art methods to demonstrate the effectiveness of our MSJE in Section~\ref{experiments} and conclude the paper in Section~\ref{conclusion}.

\section{Related Work}
\label{relatedwork}
The prevalence of social media has enabled the rapid growth of user-generated and shared online recipes and food images. There are three popular learning tasks on food datasets. (1) Food recognition from images, which evolves from kernel-based model~\cite{Joutou+ICIP2009} to DNN approaches~\cite{Kawano+2014,Yanai+2015}; (2) food recommendation~\cite{Fadhil-2018, Elsweiler+SIGIR-2017}, including ingredient identification~\cite{Chen+2016}, dietary recommendation for diabetics~\cite{Rehman+2017} and recipe popularity prediction~\cite{Sanjo+2017}; (3) Cross-modal food retrieval~\cite{Food101,Carvalho+SIGIR2018_AdaMine,JinJinChen+MM2018_AMSR,JinJinChen+MM2017_SAN,Salvador+CVPR2017_JESR}: Image to Recipe and  Recipe to Image. Food images are diverse in terms of background, ingredient composition, visual appearance and ambiguity. Early works~\cite{Jeno+SIGIR-2003,Sun-2011} circumvented this problem by annotating images to perceive their latent semantics. 

These approaches, however, are supervised and require users to annotate at least a small portion of images. The unsupervised solutions map images and recipe texts into a shared latent space in which their feature vectors can be compared in terms of vector similarity. A classic example is to employ canonical correlation analysis (CCA) for learning joint embeddings for semantic visual annotation~\cite{Rasiwasia+MM2010}. CCA utilizes global alignment to allow the mapping of different modalities, which are semantically similar, by maximizing the correlation between the input recipe and its associated image. Built on CCA, recent approaches have employed deep neural networks to cross-modal retrieval, such as DeViSE\cite{Frome+2013}, correspondence auto-encoder~\cite{Feng+MM2014}, adversarial cross-modal retrieval~\cite{Wang-2017}.

Most recent cross-modal recipe-image embedding approaches~\cite{Carvalho+SIGIR2018_AdaMine,JinJinChen+MM2018_AMSR,JinJinChen+MM2017_SAN,Salvador+CVPR2017_JESR,Hao+CVPR2019_ACME} resort to LSTM networks with word2vec and CNN networks to generate the recipe embedding and image embedding respectively. SAN~\cite{JinJinChen+MM2017_SAN} applied a stacked attention network to simultaneously locate ingredient regions in the image and learn multi-modal embedding features. JESR~\cite{Salvador+CVPR2017_JESR} proposed to learn a joint embedding between the two modalities by combining a pairwise cosine loss with a semantic regularization constraint. AMSR~\cite{JinJinChen+MM2018_AMSR} improves JESR by using hierarchical attention on the recipes and using simple triplet loss for joint embedding regulation. AdaMine~\cite{Carvalho+SIGIR2018_AdaMine} extends JESR by using a double triplet loss, which simultaneously leverages retrieval and class-guided features in the shared latent space. ACME~\cite{Hao+CVPR2019_ACME} proposed adversarial cross-modal embedding by using triplet loss with hard-sample mining~\cite{Hermans+2017}. 

\begin{figure*}[!t] 
  \centering
  \includegraphics[scale=0.25]{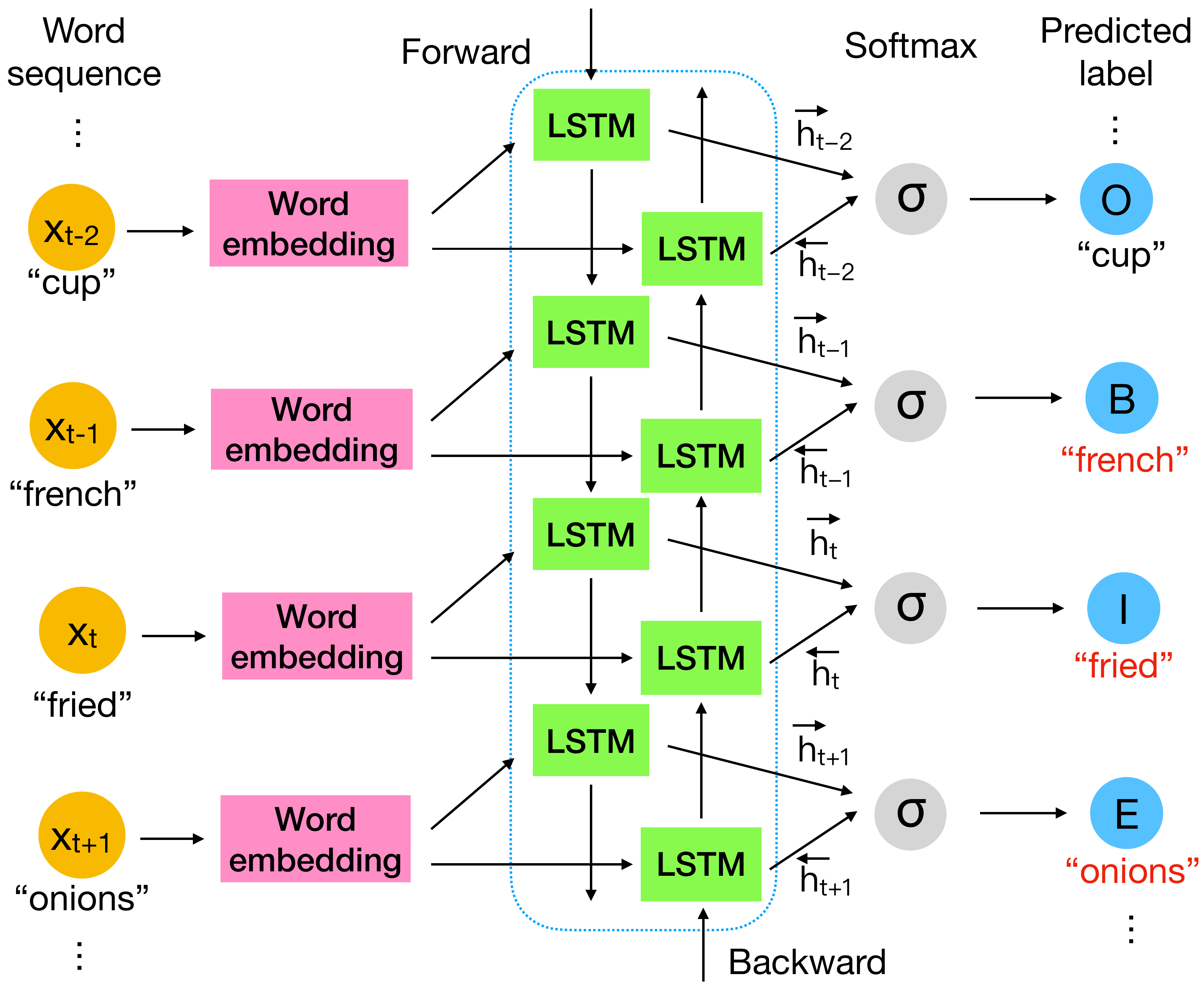}
\caption{ Using the Bi-directional LSTM to extract the ingredient names through word sentence structure analysis.  Labels "O", "B", "I" and "E" represent the word outside the target entity, the beginning position, the intermediate position, and the ending position of the entity respectively. "French fried onions" is extracted by LSTM as a true term in this example. }
  \label{lstm}
\end{figure*}

\begin{figure*}[tbp]
  \centering
  \includegraphics[scale=0.42]{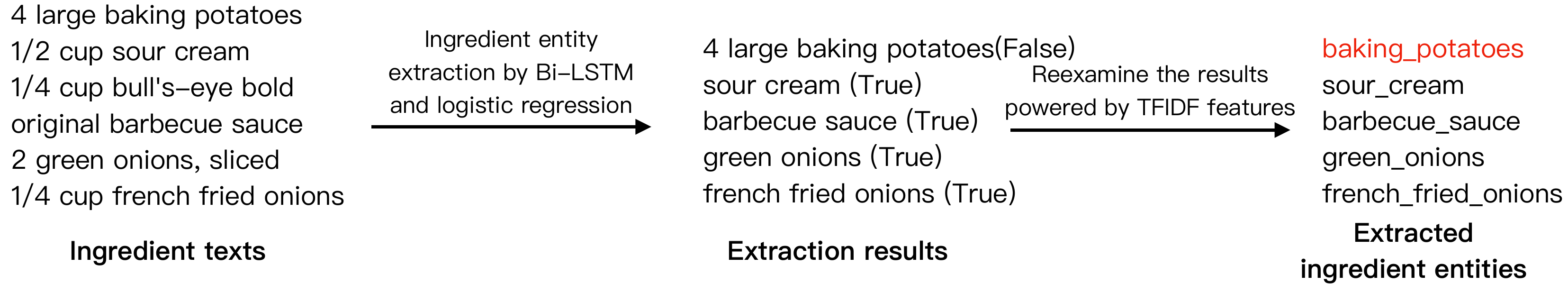}
  \caption{Reexamining the extracted word sequence marked with a false label to recover the errors from bi-directional LSTM based ingredient extractor. The ingredient entity "baking potatoes'' contained in the extracted word sequence "4 large baking potatoes'' labeled as false would be missed in other approaches but recovered in our proposed MSJE approach due to its relatively high TFIDF value (Recall Figure 5, the TFIDF value is 0.2165 and the threshold is set as 0.1 in our work).}
  \label{miss_ingr}
\end{figure*}

\section{MSJE Cross-Modal Embedding Learning}
\label{framework}

The problem of training a cross-modal retrieval model is a multi-tasking learning problem with dual goals: (1) It learns the feature maps of training data from each of the modalities, and (2) It learns two independent embeddings iteratively, each of which maps the $n$ dimensional feature vector from one modality to a common $d$ dimensional latent vector representation space ($d<n$), such that the similarity loss of the two modality embeddings is minimized through joint embedding optimization.

There are several open challenges: 

(1) Recipes have three text sections: title, ingredients, and cooking instruction that are related to the content of food images. Recall Figure~\ref{ab-example}, Title tends to give out the core phrase for the recipe instance, while ingredients give out the key terms (one or more words), and cooking instruction consists of a sequence of sentences describing the food preparation in the order of how ingredients are used in the cooking process. Not all the ingredients are descriptive of the associated food image(s), such as olive oil, salt. Some descriptions in cooking instruction are not directly relevant and thus cannot be translated to the visual appearance of food image content, such as add salt, garlic powder, preheat oven. 

(2) Even though both recipe text and food images are confined in the food domain, the pairing of a recipe and its associated image is the only explicit reference linkage of the two modalities. However, several complications may add confusion for learning joint embedding during the feature engineering from different modalities. For instance, multiple recipe instances may correspond to the same name, such as baked ziti. A food image query about baked ziti will result in more than one recipe instance with baked ziti as the recipe title, as shown in Figure~\ref{ab-example}(a). However, only one recipe instance with baked ziti as the recipe title will be the top-1 perfect match, even though the second recipe shares the same title of baked ziti and has similar ingredients and instructions. Figure~\ref{ab-example}(b) shows another complication. The two recipe instances have the same title and similar ingredients, but the food images associated with the two recipes display very different visual appearances beyond optic variations, which are mainly the results of different cooking processes due to differences in specific cooking instructions. Finally, one recipe instance may be associated with more than one image instances as shown in Figure~\ref{ab-example}(c). These and other possible complications not only add significant complexity but also pose some unique challenges when learning cross-modal neural joint embedding, which are different from those in visual question-answering~\cite{Anto+ICCC2015}, visual annotation~\cite{Smeaton+2006} and image captioning~\cite{Vinyals+CVPR2015}.

\begin{figure*}[tbp]
  \centering
  \includegraphics[scale=0.28]{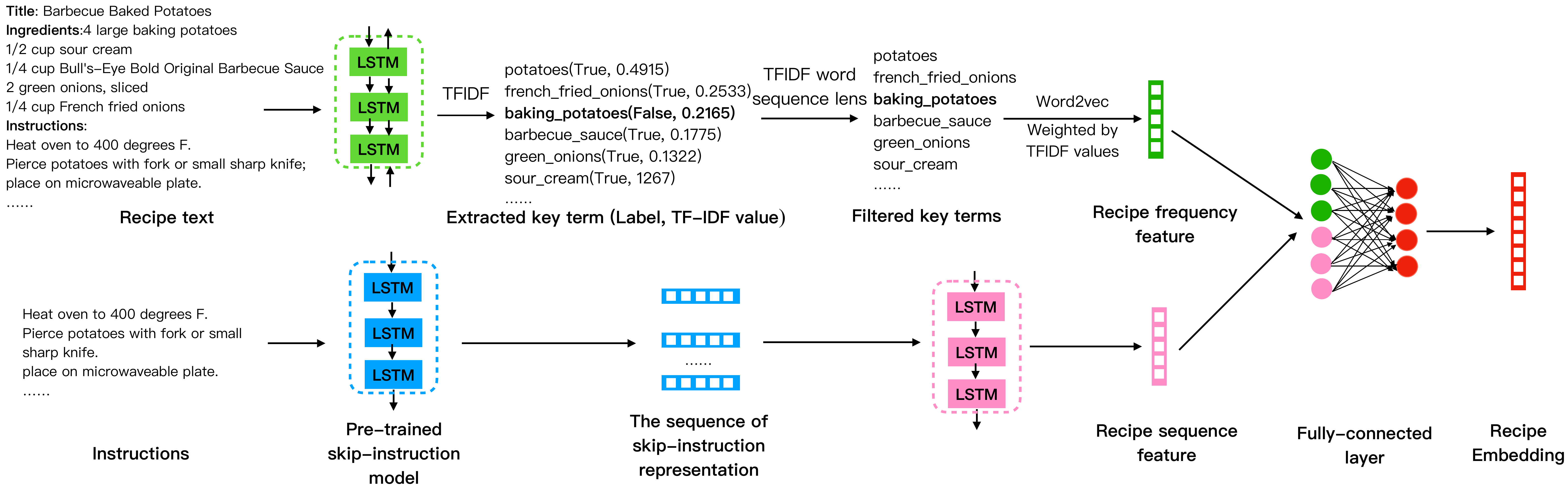}
  \caption{Semantic enhanced recipe embedding learning framework. }
  \label{sk_tfidf_recipe_branch}
\end{figure*}

\begin{figure*}[tbp]
  \centering
  \includegraphics[scale=0.28]{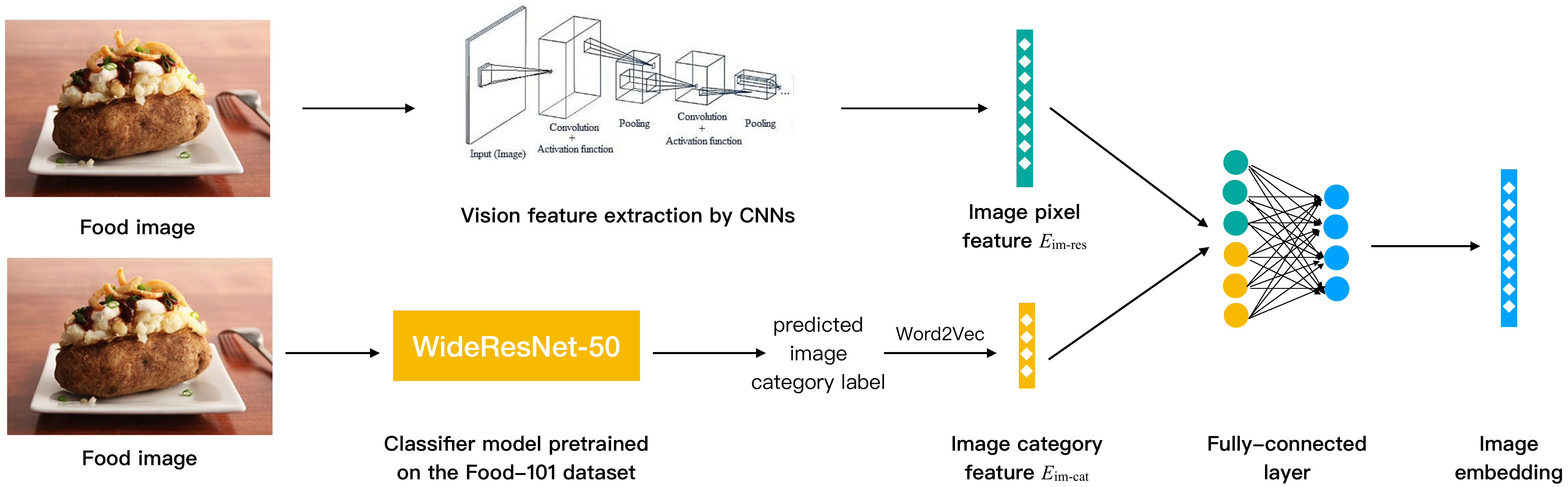}
  \caption{Semantic enhanced image embedding learning: an illustration by example}
  \label{image_branch}
\end{figure*}

\subsection{Semantics-Enhanced Recipe Embedding.} 
Learning recipe text embedding involves two subtasks: learning the text feature maps of recipes and learn the embedding of recipe feature maps to a $d$ dimensional joint latent representation space. For recipe text feature extraction, we extract the frequency semantics and sequence semantics to generate the recipe embedding. For the recipe frequency feature, MSJE analyzes word sequence and sentence structure and annotates/enhances the learned features with TFIDF semantics of main ingredients in a recipe. TF-IDF (Term Frequency-Inverse Document Frequency) is a numerical statistic score that is intended to reflect how important and how discriminative a term is with respect to a document in a given corpus (e.g., training recipe set). It is computed based on two metrics: term frequency per document and the inverse document frequency of the term over the set of documents. The recipe sequence feature is obtained by the two-stage LSTM networks~\cite{Salvador+CVPR2017_JESR}. LSTM (Long Short Term Memory networks) is a type of recurrent neural network (RNN), which is capable of learning long-term dependencies.

The recipe frequency feature mainly focuses on the significance of each ingredient entity to its recipe. Bi-directional LSTM networks and logistic regression are utilized as ingredient entity extractor~\cite{Salvador+CVPR2017_JESR}. We first use bi-directional LSTM networks to analyze the word sentence of each ingredient in the list, such as "1/4 cup of French fried onions", and learn to identify/extract only those word sequences that could be ingredients, like "French fried onions", "tomato sauce". Figure~\ref{lstm} shows an example, which uses the bi-directional LSTM networks to extract the ingredient entity from the original ingredient text.  The word sequence "French fried onions" would be extracted according to the predicted "B", "I" and "E" labels from the original ingredient text "1/4 cup French fried onions".

Then we perform binary logistic regression on the set of word sequences extracted from LSTM into true or false to adjudicate if these extracted terms are real ingredient entities or not. Finally, each true word sequence is transformed to a single term ingredient entity by connecting the words with underline, e.g., "French\_fried\_onions", "tomato\_sauce". We create an ingredient entity dictionary to record the transformation between the word sequence of an ingredient to a single term ingredient entity, like "French fried onions" $\rightarrow$ "French\_fried\_onions". Due to the accuracy limitation of the ingredient entity extractor, we revisit those false word sequences with this ingredient entity dictionary in order to keep more real ingredient entities. Figure~\ref{miss_ingr} shows an example. By reexamining those word sequences marked as "false", we correct errors of feature learning via ingredient entity extractor and recover important ingredient entities that would otherwise be missed, i.e., "Baking Potatoes" in Figure~\ref{miss_ingr}. This shows that unlike the existing approaches~\cite{Salvador+CVPR2017_JESR, Carvalho+SIGIR2018_AdaMine,Hao+CVPR2019_ACME} that rely solely on the LSTM for recipe feature learning, the proposed approach provides more robust feature extractions and can recover important ingredient entities that were missed. For the cooking instruction (and title) of a recipe, before performing similar word and sentence sequence feature learning, we first utilize the ingredient entity dictionary to transform the word sequence of an ingredient entity into its single term format. Then we analyze sentence sequence patterns and extract the key terms in cooking instruction of a recipe with TFIDF enhanced semantics.

Next, a word2vec model~\cite{Mikolov-2013} is trained over the TFIDF enhanced term corpus from all recipe texts in the training data, based on the Continuous Bags of Words (CBOW) model framework, which computes the target word probability based on its surrounding context words, and each word in the training corpus obtains its word2vec vector in the joint latent representation space. Every word in the training recipe corpus can obtain its word2vec vector through the learned word embedding matrix, which captures the word sequence and sentence sequence patterns in the text recipes.

\begin{figure*}[!t] 
  \centering
  \includegraphics[scale=0.28]{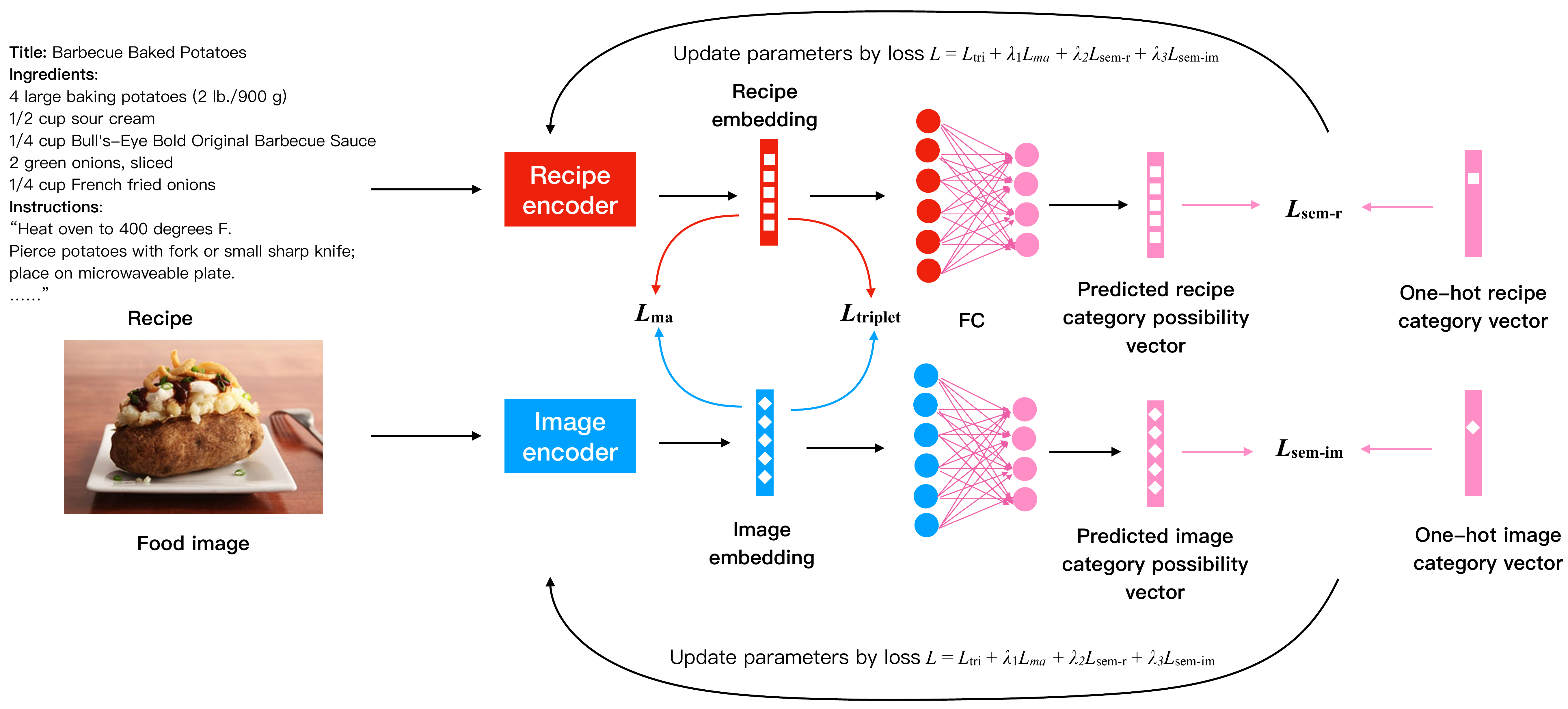}
  \caption{The framework of MSJE cross-modal embedding learning approach}
  \label{general_framework}
\end{figure*}

The upper part of Figure~\ref{sk_tfidf_recipe_branch} provides an example illustration of how to generate the recipe frequency feature. 
For each key-term $kt_i$ extracted from the recipe ($i=1,...,N_t$,  $N_t$ is the number of key terms extracted from the recipe), we compute the TFIDF value $w_i$ of each key-term. The word2vec representation $x_i$ for each key-term $kt_i$ can be obtained by the learned word2vec matrix. By combining the word2vec vectors $X=[x_1, x_2, \cdots, x_{N_t}]$ of all the key-terms from the recipe, weighted by their TFIDF values $W=[w_1,w_2,\cdots, w_{N_t}]$, we get the recipe frequency feature by $WX^T$.

For the recipe sequence feature, we extract the sequence semantics from the word and sentence sequence in cooking instructions. In order to capture different levels of granularity in the instruction text, we utilize the two-stage LSTM networks to learn and obtain the recipe sequence features. The word-level feature of each instruction sentence is obtained by the pre-trained skip-instruction model~\cite{Salvador+CVPR2017_JESR}, built upon LSTM networks and the technique of skip-thoughts~\cite{Kiro-2015}, while the sentence-level feature of the sequence of instructions is learned from scratch by another LSTM networks. Figure~\ref{sk_tfidf_recipe_branch} shows our semantics enhanced recipe embedding learning approach by combining recipe keyword (ingredient) frequency features and recipe (instruction) sequence semantics.

\subsection{ Semantics Enhanced Image Embedding}
For image embedding learning, most of the existing works resort to pre-trained CNN networks as image encoder, e.g., VGG-16~\cite{Simonyan-2014} and ResNet-50~\cite{He+CVPR2016} models. However, different recipes for a dish may have different visual appearances due to different cooking procedures and use of ingredients. We argue that the neural features learned from the pixel grid of food image should be enhanced by the semantic knowledge specific to the corresponding recipe text to align the image embedding closer to the associated recipe text embedding in the jointly learned latent representation space.  In our proposed MSJE, we incorporate the category semantics for food images based on the Food-101 dataset~\cite{Food101} as an additional dimension of recipe-image relationship semantics to strengthen the alignment between the recipe text embedding and the food image embedding. 

Figure~\ref{image_branch} provides an illustrative example. First, we train a CNN classifier using WideResNet-50~\cite{wideResnet} on the food101 dataset, which will predict one of the 101 food categories for a food image associated with a recipe. Similar to key term embedding during the recipe embedding process, we also generate the image category embedding $E_{im-cat}$ for this image category annotation using the word2vec model. We integrate this image category embedding $E_{im-cat}$ with the ResNet-50 auto-encoder generated food image embedding $E_{im-res}$ by vector concatenation operation, and then followed by a fully-connected layer to get the semantics enhanced image embedding.

\subsection{ Semantics Enhanced Joint Embedding Learning.}

Figure~\ref{general_framework} illustrates how the MSJE approach leverages the TFIDF semantics and the category semantics to optimize the cross-modal joint embedding learning over four objectives: to learn the recipe embedding and the food image embedding, to align the embeddings of the two modalities with the aim of making the feature distribution of cooking recipes and food images indistinguishable, and finally, to enforce the semantic regularization constraint such that both embeddings can be refined iteratively to perform well in the classification task.

The goal of learning semantics enhanced joint embedding is to design an objective function that can leverage multimodal semantics to enhance the optimization for minimizing the similarity loss (distance) between the recipe embedding $E_r^{rt_i}$ and the image embedding $E_{im}^{fm_i}$ for a given recipe input pair: recipe text ($rt_i$) and food image ($fm_i$). In MSJE, we utilize the batch hard triplet loss~\cite{Hermans+2017} to implement the constraint optimization for joint embedding learning:
 
\begin{equation}
L = L_{tri} + \lambda_1 L_{ma} +  \lambda_2  L_{sem-r} + \lambda_3  L_{sem-im}
\end{equation}
\noindent 
where $\lambda_1$, $\lambda_2$ and $\lambda_3$ are trade-off parameters,  which are set as 0.005, 0.005 and 0.005 respectively. $L_{tri}$ is the loss associated with the retrieval task over instance-based triplets $(x^a, x^p, x^n)$. $ L_{ma}$ is the adversarial loss to align the embeddings from two modalities. $ L_{sem-r}$ and $L_{sem-im}$ are the losses over the recipe embedding $E_{r}$ and the image embedding $E_{im}$ respectively. 

For the batch-hard triplet loss, a triplet $(x^a, x^p, x^n)$ is utilized to define the loss function: $x^a$ represents a feature embedding as an anchor point in one modality, $x^p$ denotes a positive feature embedding and $x^n$ denotes a negative feature embedding from the other modality. The optimization is to ensure that the positive instance is the one that should be close to the anchor point, and the negative instance should be far away from the anchor point. In the design of our MSJE, we consider two types of triplets: (1) using the recipe embedding as the anchor $E_{r}^a$, and (2) using the image embedding as the anchor $E_{im}^a$. In the case of using batch hard triplet loss, for a given anchor, we choose the most distant positive instance and the closest negative instance during the training process. The batch hard triplet loss is given as:
\begin{small}
\begin{equation}
\begin{aligned}
L_{tri} =\sum_{\substack{i=1\\j=1}}^B[\max{d(E_{r}^{i,a}, E_{im}^{i,p})}-\min_{\substack{i\neq j}}{d(E_{r}^{i,a}, E_{im}^{j,n})} + m_2]_+ \\ + \sum_{\substack{i=1\\j=1}}^B[\max d(E_{im}^{i,a}, E_{r}^{i,p})-\min_{\substack{i\neq j}}d(E_{im}^{i,a}, E_{r}^{j,n}) + m_2]_+ \\ 
\end{aligned}
\label{tri_loss}
\end{equation}
\end{small}
\noindent 
Here $d(\cdot)$ is the euclidean distance, $B$ is the number of recipe-image pairs in one batch, superscripts $a$, $p$ and $n$ refer to anchor, positive and negative instances respectively, superscripts $i$, $j$ refer to the $i$-th and $j$-th recipe-image pair, and $\alpha$ is the margin of error. 

The second objective is to align the embeddings from two modalities to deal with the possibility of poor generalization and slow convergence due to the different distributions of encoded features. Unlike ACME in~\cite{Hao+CVPR2019_ACME}, our design of this modality alignment loss aims to align the distributions of ultimate layer features, instead of the penultimate layer features. 
By utilizing the competing strategy in GAN, we can view the modality alignment loss as an adversarial loss because when given an embedding from one modality, a discriminator $D_M$ cannot distinguish whether the feature embedding is generated from the food image or recipe text.
Hence, one can make the distributions of the feature embeddings from two modalities more similar with the objective $L_{ma}$ calculated as follows:

\begin{equation}
\begin{aligned}
L_{ma} = \mathbb{E}_{x_{im} \sim P_{im}}[\log D_M(\phi_{im}(x_{im}))]_+ \\+ \mathbb{E}_{x_r \sim P_r}[\log (1-D_M(\phi_r(x_r)))]_+
\end{aligned}
\label{ma_loss}
\end{equation}
\noindent where $\phi_{im}$ and $\phi_{r}$ is the function to encode the raw image feature $x_{im}$ and recipe feature $x_r$ into image embedding and recipe embedding respectively. $P_{im}$ and $P_{r}$ are the data distributions of food images and cooking recipes.

Finally, for each recipe (text and image pair), we use the learned image embedding and the learned recipe embedding to produce a fully-connected layer. First, this joint fully-connected layer will be fed into the cross-modal embedding learning with equal contribution to the training of joint embedding (can be seen as using the equal weight). Second, we also leverage this joint fully-connected layer to perform the classification task, which assigns a proper categorization label to each input recipe (text and image). This learned categorization label provides additional semantics to the optimizations in joint embedding loss functions. In MSJE, we utilize this semantically enhanced category generation module to address the inherent problems of semantic regularization in JESR~\cite{Salvador+CVPR2017_JESR}, such as the large number of recipe input under background category and mismatching between recipe text and its associate food image. Our categorization module can effectively categorize the input recipe (text and image pair) by integrating bigram-based recipe text categorization with the Food101 image categorization to ensure every input recipe (text and image pair) will be assigned a meaningful classification label instead of a background label. To ensure that the learned recipe embedding and image embedding contain the same category semantics during the joint embedding learning process, we keep the semantic regularization constraint on the two learned embeddings to facilitate the objective of minimizing the distance between matched recipe-image pairs and maximizing (enlarging) the distance of unmatched and irrelevant pairs, formulated as:

\begin{equation}
\label{sem_r}
L_{sem-r} = -\frac{1}{N}\sum^{N_c}_{i=1}{y_i^r \log(\hat{y}_i^r)}
\end{equation}

\begin{equation}
\label{sem_im}
L_{sem-im} = -\frac{1}{N}\sum^{N_c}_{i=1}{y_i^{im} \log(\hat{y}_i^{im})}
\end{equation}
\noindent
where $L_{sem-r}$ is the loss of semantic regularization on the recipe branch, while $L_{sem-im}$ is on the image branch. $N$ is the number of the different recipe-image pairs, $N_c$ is the number of category labels. $y_i^r$ and $\hat{y}_i^r$ are the true and estimated possibilities that the recipe embedding belongs to the $i^{th}$ category label, while $y_i^{im}$ and $\hat{y}_i^{im}$ are for image embedding.

Note that this semantic regularization adds an extra layer on the learned embeddings to perform the food classification, aiming for regularizing the latent space embeddings. There are two alternative designs for the category assignment algorithm: From the released codes of JESR~\cite{Salvador+CVPR2017_JESR}, its category assignment algorithm uses Food 101 as a secondary source for category alignment by performing the following three tasks: (1) Obtain the top 2000 most frequent bigrams in recipe titles from the training set and manually remove those that do not have discriminative food properties (e.g., 5 minutes and \& !, enjoy!). Assign the bigrams as the category to all recipes whose title contains the bigram.
(2) The category labels of Food-101 dataset are assigned to the remaining recipes if their recipe titles contain the category label. 
(3) The remaining recipes whose titles contain neither one of the filtered bigrams nor one of the food101 categories are assigned to the background category. 

Unfortunately, this approach assigns nearly half of the recipes with a "background" category label, due to the lack of matching to a real category label in JSER, which limits the effectiveness of JSER. 
In designing our semantic regulation algorithm for MSJE, we use the Food 101 categories as the primary source of recipe category labels and use bigram categories as the second source, followed by examining ingredients and cooking instructions for matching with the above two category sources. Further, we use the image classifier of Food 101 on the image associated with the recipe to predict and assign the top-1 category label to the recipe.  
(1) For a recipe whose title contains a category label in the food101 categories, we first assign this category to the recipe. Note that if the recipe title is "meatloaf" or "sweet cookies", then it can be tagged with "meat loaf" or "sweet cookie" category label. 
(2) To make a fair comparison, in the prototype of MSJE, we also obtain the top 2000 most frequent bigrams in recipe titles of the training set in the same fashion as JESR and remove more inappropriate ones (e.g. home made, new style and oven fried). For those recipes that do not contain any of the category labels from Food 101, if a recipe title contains one of the filtered bigrams, then the bigram is assigned to this recipe as the category label. 
(3) For each of the remaining recipes, If the ingredients or the cooking instruction of this recipe contain one of the Food 101 or bigram categories, then this recipe will be assigned to the category.
(4) All the remaining recipes are assigned with one of Food-101 categories based on the top 1 category label predicted by the food image classifier. 

In summary, our cross-modal training framework MSJE incorporates different semantics for each modality at different learning stages of training the cross-modal joint embedding model, including extracting TFIDF features from title, ingredients and cooking instructions of recipes and incorporating them in LSTM based feature engineering, word2vec based auto-encoder for recipe feature mapping, CNN based auto-encoder for food image feature projection, as well as optimizing and regulating the loss function during iterative learning of cross-modal joint embedding. In the next section, we report our experimental evaluation on the effectiveness of MSJE design compared to the state-of-the-art approaches for both image-to-recipe and recipe-to-image queries on the benchmark Recipe1M dataset.

\newcommand{\tabincell}[2]{\begin{tabular}{@{}#1@{}}#2\end{tabular}}  

\begin{table*}[!t] 
		\center
		\caption{ Performance comparison of our approach (MSJE) with existing state of the art methods. The symbol "-" indicates that the results are not available from the corresponding method.}
		\label{main_results}
		\begin{tabular}{cc|cccc|cccc}
		\hline
		\multirow{2} * {\tabincell{c}{Size of \\test-set}} & \multirow{2} * {Approaches}  &\multicolumn{4}{c}{Image to recipe retrieval} & \multicolumn{4}{c}{Recipe to image retrieval } \\
		 \cline{3-10}
		   ~ & ~ & MedR$\downarrow$ & R@1$\uparrow$ &R@5$\uparrow$ & R@10$\uparrow$ & MedR$\downarrow$ & R@1$\uparrow$ &R@5$\uparrow$ & R@10$\uparrow$ \\  
		\hline
		\multirow{10} *{1k}& CCA~\cite{CCA} & 15.7 & 14.0 & 32.0 & 43.0 & 24.8 & 9.0 & 24.0 & 35.0\\
		~ & SAN~\cite{JinJinChen+MM2017_SAN}&16.1 & 12.5 & 31.1 & 42.3 & - & - & - & - \\		
		~ & JESR~\cite{Salvador+CVPR2017_JESR} & 5.2 & 24.0 & 51.0 & 65.0 & 5.1 & 25.0 & 52.0 & 65.0 \\
		~ & Img2img+JESR~\cite{lien2020recipe}  & - & - & - & - & 5.1 & 23.9 & 51.3 & 64.1 \\ 
		~ & AMSR~\cite{JinJinChen+MM2018_AMSR}  & 4.6 & 25.6 & 53.7 & 66.9 & 4.6 & 25.7 & 53.9 & 67.1 \\
		~ & AdaMine~\cite{Carvalho+SIGIR2018_AdaMine}  & \textbf{1.0} & 39.8 & 69.0 & 77.7 & \textbf{1.0} & 40.2 & 68.1 & 78.7 \\
		~ & R$^2$GAN~\cite{zhu2019r2gan} &2.0 & 39.1 & 71.0 & 81.7 & 2.0 & 40.6 & 72.6 & 83.3 \\
		~ & MCEN~\cite{fu2020mcen} &2.0 & 48.2 & 75.8 & 83.6 & 1.9 & 48.4 & 76.1 & 83.7 \\
		~ & ACME~\cite{Hao+CVPR2019_ACME} & \textbf{1.0} & 51.8 & 80.2 & 87.5 & \textbf{1.0} & 52.8 & 80.2 & 87.6 \\
		~ & \textbf{MSJE} & \textbf{1.0} & \textbf{56.5}  & \textbf{84.7} & \textbf{90.9} & \textbf{1.0}  & \textbf{56.2}  & \textbf{84.9}  & \textbf{91.1}  \\
		\hline
		 \multirow{8} *{10k} & JESR~\cite{Salvador+CVPR2017_JESR} & 41.9 & - & - & - & 39.2 & - & - & - \\
		~ & AMSR~\cite{JinJinChen+MM2018_AMSR} & 39.8 & 7.2 & 19.2 & 27.6 & 38.1 & 7.0 & 19.4 & 27.8 \\
		~ & AdaMine~\cite{Carvalho+SIGIR2018_AdaMine}  & 13.2 & 14.9 & 35.3 & 45.2 & 12.2 & 14.8 & 34.6 & 46.1 \\
		~ & R$^2$GAN~\cite{zhu2019r2gan} &13.9 & 13.5 & 33.5 & 44.9 & 12.6 & 14.2 & 35.0 & 46.8 \\
		~ & MCEN~\cite{fu2020mcen} & 7.2 & 20.3 & 43.3 & 54.4 & 6.6 & 21.4 & 44.3 & 55.2 \\
		~ & ACME~\cite{Hao+CVPR2019_ACME} & 6.7 & 22.9 & 46.8 & 57.9 & 6.0 & 24.4 & 47.9 & 59.0 \\
		~ & Triplet+BOW+ResNet~\cite{Fain+2019} & 5.9 & 24.4 & 49.4 & 60.5 & - & - & - & - \\
		~ & \textbf{MSJE} & \textbf{5.0} & \textbf{25.6}  & \textbf{52.1} & \textbf{63.8} & \textbf{5.0}  & \textbf{26.2}  & \textbf{52.5}  & \textbf{64.1}  \\
		\hline
		\end{tabular}
\end{table*}

\section{Experiments}
\label{experiments}
\subsection{Dataset and Evaluation Metrics}
We evaluate the effectiveness of our proposed approach on Recipe1M dataset~\cite{Salvador+CVPR2017_JESR}, which was compiled from dozens of popular cooking websites. Each recipe has 1-5 associated images but every image only belongs to one recipe, and 51.4\% of image-recipe pairs in the dataset have one of the 1047 category labels generated from category labels of Food-101 dataset and the frequent bigrams of recipe titles, while the rest of image-recipe pairs are categorized into a "background" category. Following preprocessing in~\cite{Salvador+CVPR2017_JESR}, duplicate recipes, recipes without images, the unreadable recipes without any nouns or verbs are filtered out, resulting in 238,399 matching pairs of images and recipes for the training set, and 51,116 and 51,303 matching pairs for validation and test respectively. 

Experiment setups follow the same guidelines as the prior works ~\cite{Salvador+CVPR2017_JESR, JinJinChen+MM2018_AMSR,Carvalho+SIGIR2018_AdaMine,Hao+CVPR2019_ACME}. Specifically, we first sample 10 unique subsets of 1,000 or 10,000 matching recipe-image pairs from the test set. Then we consider each item in one modality as a query (e.g., an image), and rank instances in the other modality (e.g., recipes) using the Euclidean distance between the query embedding and each candidate embedding from the other modality in the test set. We calculate the median retrieval rank (MedR) and the recall percentage at top K (R@K), i.e., the percentage of queries for which the matching answer is included in the top K results.

\subsection{Implementation Details}
Public word2vec toolkit~\footnote{https://code.google.com/archive/p/word2vec/} is used to train the Continuous Bag-of-Words (CBOW) version of word2vec model~\cite{word2vec-2013} on the corpus of recipe texts in the Recipe1M dataset. The dimensions of joint embedding and word2vec embedding are set as 1024 and 300 respectively. Adam optimizer~\cite{Adam-2014} is employed for model training with the initial learning rate set as $10^{-4}$ in all experiments, with the mini bath size of 100. In each mini-batch, every recipe (image) is restricted to have exactly one ground-truth image (recipe), and the remaining images (recipes) are regarded as negative instances during training. The deep neural networks are implemented on the Pytorch platform and trained on a single NVidia Titan X Pascal server with 12GB of memory. The training phase for MSJE joint embedding on Recipe1M dataset requires about 13 hours to generate the trained model, meeting the convergence condition of the training. When using the trained cross-modal embedding model to perform recipe-image cross-modal retrieval, it takes only about 0.13 seconds to perform a cross-modal retrieval on the 1k test set, and about 1.5 seconds on the 10k test set. The model size for the MSJE trained recipe-image embedding model is 434.98MB (including the learned discriminator networks). In comparison, the model size of the JESR trained model is 540.85MB, and the model size for the ACME trained model is 376.11MB.

\begin{figure*}[!t] 
  \centering
  \includegraphics[scale=0.27]{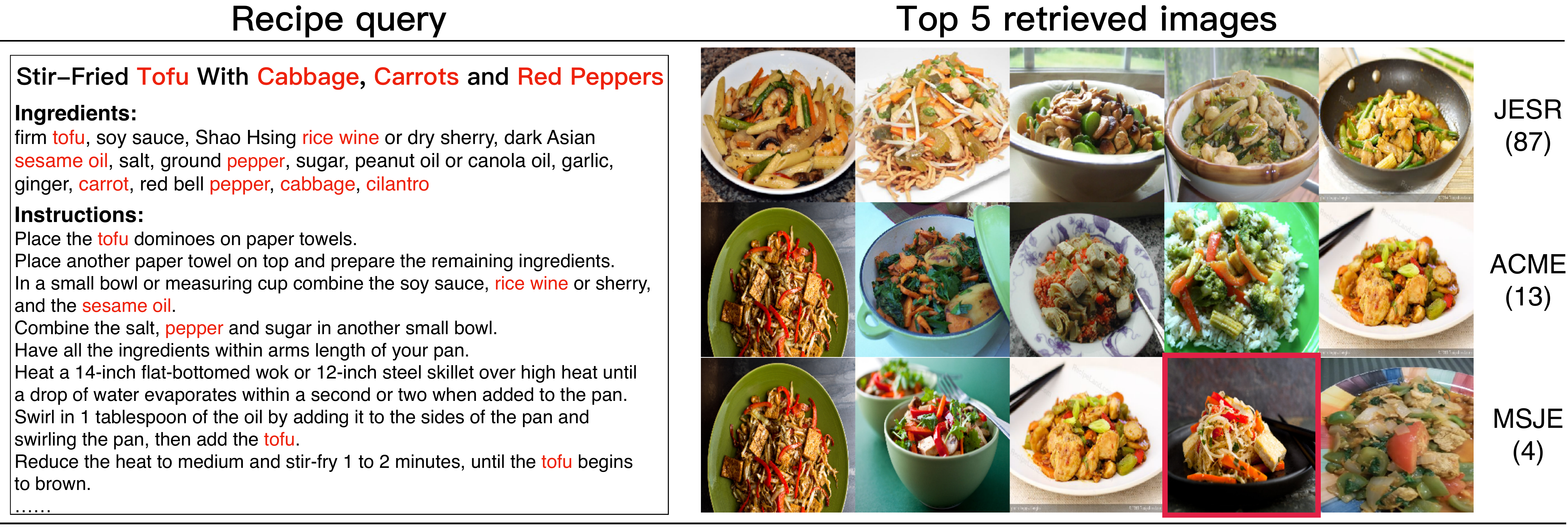}
  \caption{ The visual results of recipe-to-image retrieval on the 10k test set by using JESR, ACME and our MSJE approaches. The matched images are marked with a solid red box. Words in red are the key-terms captured by TFIDF-enhanced recipe embedding in MSJE. We list the position where the matched image is retrieved under the approach name.}
  \label{msje-com-r2i}
\end{figure*}

\begin{figure*}[!t] 
  \centering
  \includegraphics[scale=0.27]{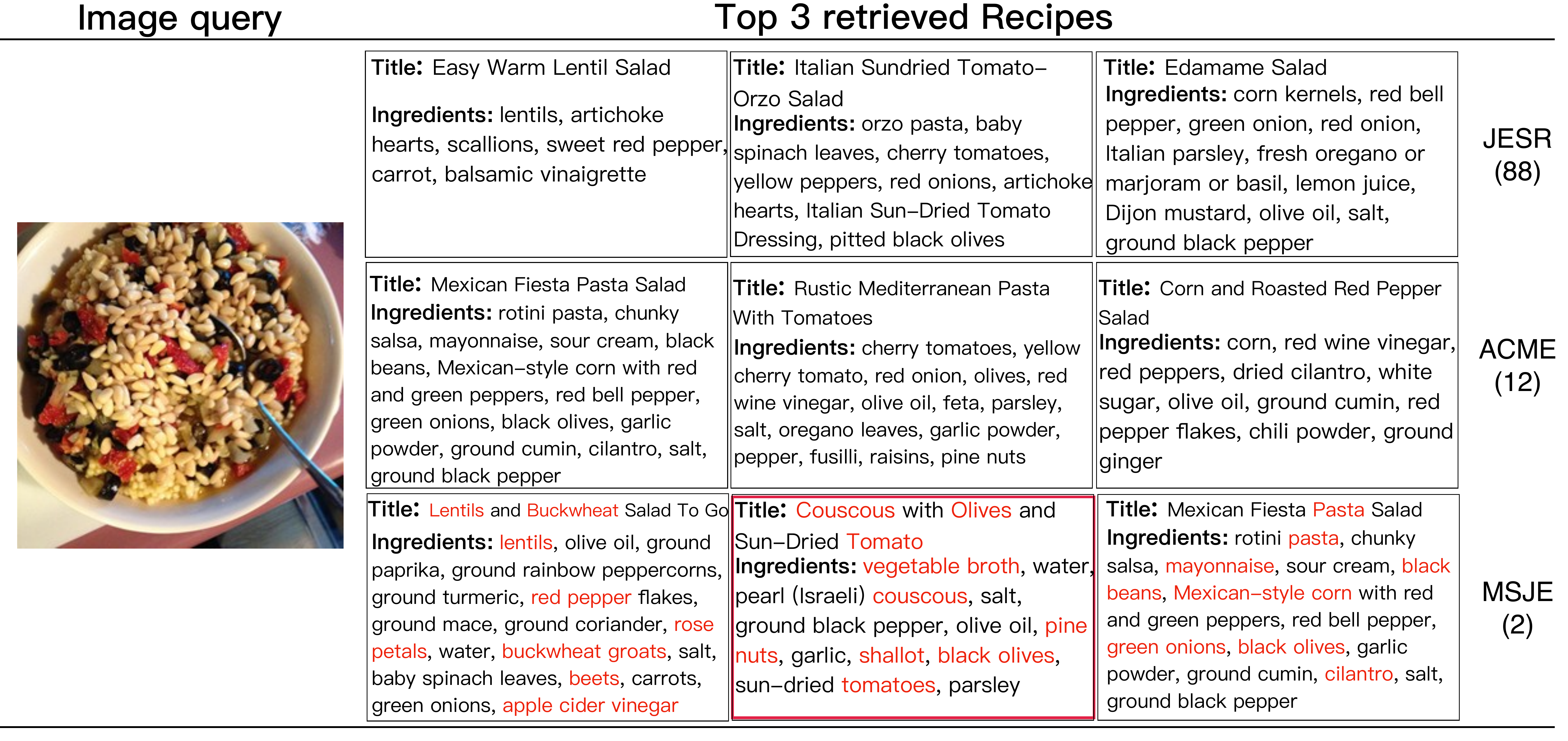}
  \caption{The visual results of image-to-recipe retrieval on the 10k test set by using JESR, ACME and our MSJE approach. The matched recipes are marked with a solid red box. Words in red are the key-terms captured by TFIDF-enhanced recipe embedding in MSJE. We list the position where the matched image is retrieved under the approach name.}
  \label{msje-com-i2r}
\end{figure*}

\subsection{Baselines}

 We consider the following state-of-the-art methods as the baselines in evaluation: 
 
\textbf{Canonical Correlation Analysis (CCA)~\cite{CCA}}: CCA learns two linear projections for mapping text and image features to a common space that maximizes feature correlation. 

\textbf{Stacked Attention Networks (SAN)~\cite{JinJinChen+MM2017_SAN}}: SAN applied a stacked attention network to simultaneously locate ingredient regions in the image and learn multi-modal embedding features between ingredient features and image features through a two-layer attention mechanism.

\textbf{Joint Embedding with Semantic Regularization (JESR) ~\cite{Salvador+CVPR2017_JESR}}: JESR obtains the recipe representation using LSTM networks on the ingredients and instructions text, the image representation using ResNet-50, and train a joint embedding of the two modalities using pairwise cosine loss with a regularization term to penalize the learned embedding features if they fail in performing food categorization.

\textbf{Recipe Retrieval with visual query of ingredients (Img2img+JESR) ~\cite{lien2020recipe}}: Based on the framework of JESR, Img2img+JESR appends the representation of the ingredient image set to the concatenation of ingredient embedding and instruction embedding, and then the concatenation of three embeddings will be transformed into one joint recipe embedding through a fully-connected layer.

\textbf{Attention Mechanism with Semantic Regularization (AMSR) ~\cite{JinJinChen+MM2018_AMSR}}: AMSR utilizes GRU networks and attention mechanism to encode recipe text at different levels (title, ingredients and instructions), and adopts the triplet loss rather than the pairwise loss to train the model.

\textbf{ADAptive MINing Embedding (AdaMine) ~\cite{Carvalho+SIGIR2018_AdaMine}}: AdaMine utilizes a double triplet loss where the two triplets are built on the matching recipe-image relationship and their category labels, with the adaptive learning integrated into their triplet loss strategy.

\textbf{Recipe Retrieval with GAN (R$^2$GAN)~\cite{zhu2019r2gan}}: R$^2$GAN is built upon the embedding learning framework of JESR, the generative adversarial network with multiple generators and discriminators and two-level ranking loss to learn compatible embeddings for cross-modal similarity measure.

\textbf{Modality-Consistent Embedding Network (MCEN)~\cite{fu2020mcen}}: MCEN exploits the latent alignment with cross-modal attention mechanism and shares the cross-modal information to the joint embedding space with stochastic latent variable models.

\textbf{Adversarial Cross-Modal Embedding (ACME)~\cite{Hao+CVPR2019_ACME}}: ACME uses the triplet loss empowered with hard sample mining and an adversarial loss to align the embeddings from the two modalities. Besides, they try to utilize the cross-modal translation consistency component to save the information lost in the training process.

\textbf{Triplet+BOW+ResNet~\cite{Fain+2019}}: Triplet+BOW+ResNet pre-trains a classifier based on the standard BOW model with category labels extracted by themselves. The average of word embeddings for the recipe text is used as recipe embedding. The results quoted in this paper is from the model trained on the same training dataset as ours.

\subsection{Performance Comparison}

We evaluate the performance of our proposed MSJE approach for image-to-recipe and recipe-to-image retrieval against the five baselines in Table~\ref{main_results}. The results of CCA and JESR are quoted from ~\cite{Salvador+CVPR2017_JESR}, the results of SAN and AMSR are quoted from ~\cite{JinJinChen+MM2018_AMSR}, the results of Img2img+JESR are quoted from ~\cite{lien2020recipe}, the results of AdaMine are quoted from ~\cite{Carvalho+SIGIR2018_AdaMine}, the results of R$^2$GAN are quoted from ~\cite{zhu2019r2gan}, the results of MCEN are quoted from ~\cite{fu2020mcen}, the results of ACME are quoted from ~\cite{Hao+CVPR2019_ACME} and the results of Triplet+BOW+ResNet is from ~\cite{Fain+2019}. We make four observations.

(1) Our approach has consistently high Recall@K (K=1,5,10) for both image2recipe and recipe2image queries, compared to all baselines on 1k and 10k test data, showing the effectiveness of incorporating TFIDF semantics and sequence semantics in learning cross-modal joint embedding.

\begin{table*}[t]
		\center
		\caption{Evaluation of different semantic components used in our MSJE framework.
}
		\label{ablation results}
		\begin{tabular}{c|cccc|cccc}
		\hline
		 \multirow{2} * {Component} & \multicolumn{4}{c}{Image to recipe retrieval } & \multicolumn{4}{c}{Recipe to image retrieval } \\
		 \cline{2-9}
		   ~ & MedR & R@1 &R@5 & R@10 & MedR & R@1 &R@5 & R@10 \\  
		\hline
		MSJE-b & 4.1 & 25.9 & 56.4 & 70.1 & 4.1 & 26.0 & 56.6 & 70.3\\
		\hline
		MSJE-b+SE$_{V}$& 3.4 & 28.1 & 59.5 & 73.1 & 3.0 & 29.4 & 61.2 & 74.4  \\
		MSJE-b+SE$_R$& 3.0 & 29.4 & 60.0 & 73.4 & 3.0 & 31.0 & 61.0 & 73.9 \\		
		MSJE-b+SE & 3.0 & 30.5 & 61.6 & 75.2 & 2.8 & 32.1 & 62.6 & 75.4\\
		\hline
		MSJE-b+SE+OSR & 2.3 & 33.5 & 68.4 & 80.9 & 2.3 & 34.1 & 68.3 & 81.4\\
		MSJE-b+SE+MA & 2.5 & 36.0 & 65.2 & 77.3 & 2.5 & 36.0 & 65.2 & 77.8\\
		MSJE-b+SE+TRI & 2.0 & 46.5 & 78.1 & 86.7 & 2.0 & 47.1 & 78.3 & 86.8 \\ 
		MSJE-b+SE+JELO (MSJE) & \textbf{1.0}  & \textbf{56.5}  & \textbf{84.7} & \textbf{90.9}& \textbf{1.0}  & \textbf{56.2}  & \textbf{84.9}  & \textbf{91.1}\\  	
		\hline
		\end{tabular}
\end{table*}

\begin{figure*}[htbp] 
  \centering
  \includegraphics[scale=0.27]{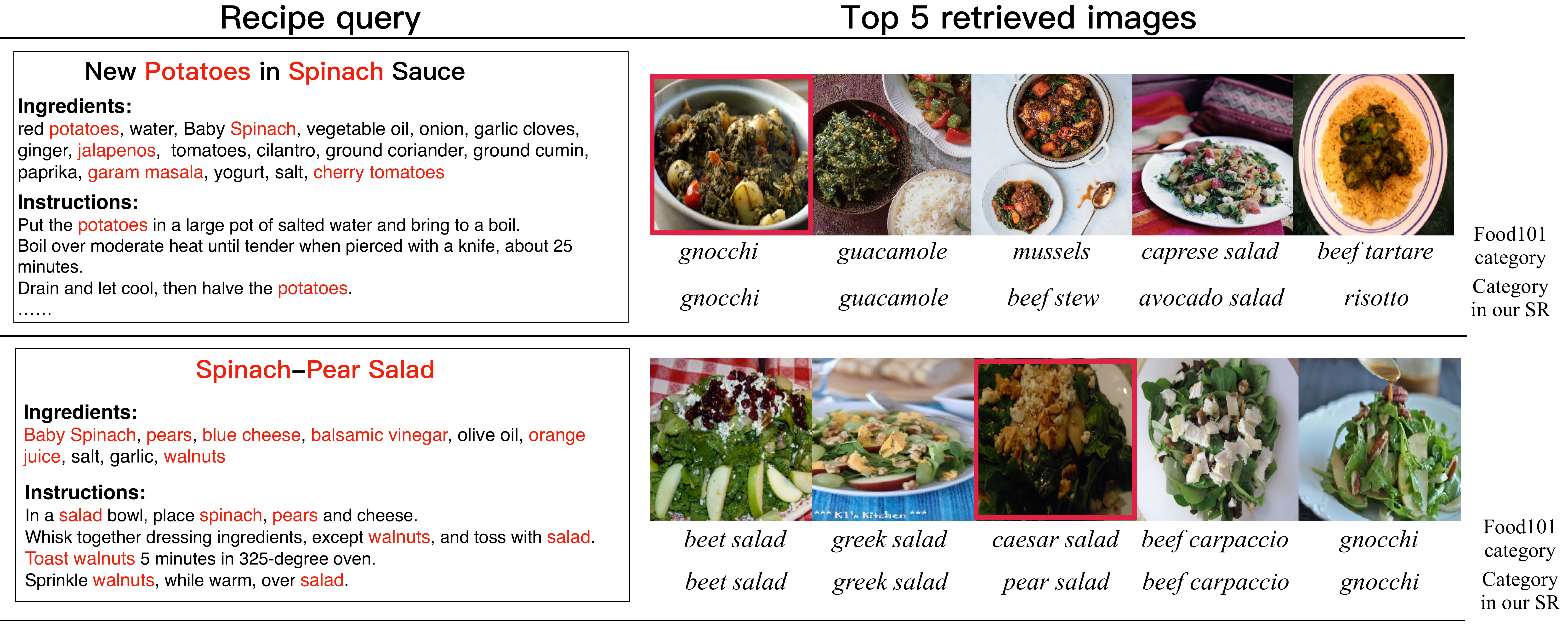}
  \caption{ The visual results of recipe-to-image retrieval on the 10k test set by using our MSJE approach. The food101 category used in the image embedding and the category used in our semantics regularization (denoted as category in our SR) are listed under the retrieved food images. The matched images are marked with a solid red box. Words in red are the key-terms captured by TFIDF-enhanced recipe embedding in MSJE.}
  \label{msje-r2i}
\end{figure*}

(2) SAN and the global alignment model CCA are less effective than other models, because (a) other models utilize the deep neural networks for cross-modal joint embedding learning and (b) even though SAN adopts the stacked networks, it learns the recipe text embedding only by extracting recipe features from the ingredient list and it uses VGG-16 for image feature learning, compared to ResNet-50 in other baseline approaches. 

(3) JESR is the first to show the use of LSTM networks to encode the ingredients and instructions for the recipe embedding and to add a semantic regularization term, improving CCA and SAN by a substantial margin. 

(4) Img2img+JESR, AMSR, AdaMine, R$^2$GAN, MCEN and ACME enhance JESR by incorporating more data or more semantics in feature engineering and joint embedding learning. One of the main reasons that the proposed MSJE approach outperforms all baselines is to incorporate TFIDF semantics, sequence semantics and category semantics in extracting recipe features, learning recipe embedding, enhancing image embedding learning, and enforcing semantics enhanced regulation and optimization for the cross-modal joint embedding learning.

Figure~\ref{msje-com-r2i} provides visualization on one recipe-to-image query using JESR, ACME and our MSJE. Obviously, JESR and ACME both fail to retrieve the matched food image in the top 5 results, but our MSJE can retrieve the associated image at the top 4 place. This is because our MSJE approach incorporates the TFIDF recipe semantics and successfully captures the discriminative ingredients (e.g. "tofu" and "red pepper" in this example). And almost all the images in the MSJE top 5 results contain the "tofu" and "red pepper" ingredients, while "tofu" is not easy to find in the top 5 results of JESR and ACME. Figure~\ref{msje-com-i2r} visualizes en image-to-recipe queries. Using our proposed MSJE approach, the matched recipe for the image query will be included in the top 3 recipes although it did not make it to the top-1 result. However, using JESR and ACME methods, the matched recipe will not be included in the top-5 retrieved recipes, showing the effectiveness of incorporating TFIDF semantics and category semantics in our semantics enhanced cross-modal embedding learning.

\subsection{Ablation studies}
Here we report the design and the results of our ablation studies, which evaluate the contributions of the different components in our MSJE approach. Specifically, MSJE-b denotes a primitive baseline framework of MSJE using the simple batch-all triplet loss, without any other semantic components proposed in this paper. We then add the main semantic components proposed in this paper to the baseline framework MSJE-b one at a time. Concretely, we first evaluate the impact of the semantics enhanced image (MSJE-b +SE$_V$) and the impact of the semantics enhanced recipe embedding learning (MSJE-b +SE$_R$) separately. Then we study the impact of incorporating both semantics enhanced image and recipe embedding learning, denoted by MSJE-b+SE. 

Next, we evaluate the impact of semantic components utilized during the joint embedding learning of our MSJE framework. Three joint embedding loss optimizations are employed in our MSJE framework: (1) the optimized semantic regularization (OSR) on both recipe text embedding and food image embedding as the secondary loss regularization; (2) the modality alignment (MA) as another secondary loss regularization, which aligns the embeddings from two modalities to improve the loss regularization in the presence of poor generalization and slow convergence during the joint embedding learning due to different distributions of the encoded features; and (3) the batch-hard triplet loss (TRI) with both recipe text anchor and food image anchor as the primary loss function to replace the batch-all triplet loss in MSJE-b. We first evaluate each of these three semantic optimizations independently, denoted by MSJE-b+SE+OSR, MSJE-b+SE+MA, and MSJE-b+SE+TRI. Then we evaluate the impact of combining all three joint embedding loss optimizations (JELO), denoted by MSJE-b+SE+JELO. 

Table~\ref{ablation results} shows the results of the ablation studies on MSJE. We make two important observations. First, each semantic component in our MSJE approach provides a positive influence towards boosting the performance of the joint embedding learning, Second, by combining all of them in concert, our MSJE approach can further improve the overall performance of joint embedding than utilizing only a subset of these semantic optimizations. This ablation study further demonstrates the novelty of our MSJE approach in delivering high-performance joint embedding learning model for recipe-image cross-modal retrieval services.

Figure~\ref{msje-r2i} provides visualization on two recipe-to-image queries using MSJE. For the first query recipe, the correctly matched food image is the top-1 result thanks to the dual benefits of semantics-enhanced joint embedding. (1) The key-term "potatoes" and "spinach" are extracted from the recipe text with TFIDF enhanced recipe feature learning. (2) The category "gnocchi" used in semantic regularization is captured in our category assignment algorithm, the food-101 category "gnocchi" is incorporated into the image embedding, and the semantics enhanced joint embedding learning is able to leverage these enhanced semantics to discriminate the matched food images from the other candidate images. Similarly, for the other recipe query, all the retrieved images are visually similar to the matched images (e.g., pears can be seen visually in the top 1 retrieved image for the second recipe query) and the assigned categories from the embeddings of both modalities are almost the same. Though our approach succeeds in the R@5 performance but failed to achieve R@1, while most of the baselines fail to include the matched images in R@5.

\section{Discussion}
{\bf Potential Application to Other Non-Food Domains.} Even though the proposed MSJE approach is designed and implemented on recipe text and food image for efficient cross-modal retrieval services, the overall design framework is applicable to solving general purpose text-image cross-modal retrieval problems in other non-food domains, such as medical procedure or health diagnosis with its associate image. In particular, the end-to-end workflow of text-image cross-modal embedding learning using MSJE, the suite of semantics based optimizations developed for learning text embedding, learning image embedding, and learning text-image cross-modal joint embedding iteratively with different loss optimizations can be easily transferred to other non-food domain. 

The MSJE framework benefits from the structure of the recipe in terms of title, ingredient list and cooking instruction. Most of the medical procedures or medical diagnoses have a similar structured layout, such as procedure title, the list of key medical equipment used in the procedure, and the instruction for the procedure. Hence, the application of MSJE approach to the collection of medical procedures and their associated images can be more straightforward than some other hard cases, such as news datasets or some social network datasets. 

The main challenge for news dataset and some social network datasets, like amazon users' reviews and product images can be summarized as follows: First, when the text datasets are less well defined with respect to the images associated with those text components, it will make the cross embedding learning more challenging. Consider a News blog about a car accident but the associated image may be centered on the people injured in the presence of the car accident. Hence, the cross-modal joint embedding learning will require additional and richer semantic feature extraction, compared to recipe and food image datasets. We plan to make the open-source of our MSJE approach on GitHub to allow domain scientists with proprietary datasets to deploy our MSJE approach, such as medical procedures with associated images, for developing text-image cross-modal retrieval services.

\section{Conclusion}
\label{conclusion}
We have presented MSJE, a multi-modal TFIDF semantics enhanced joint embedding learning approach for providing high performance cross-modality recipe retrieval tasks. This paper makes three unique contributions. First, our MSJE approach can extract and incorporate the TFIDF features in all three sections of a recipe (title, ingredients and cooking instruction). Second, our MSJE approach combines TFIDF features with sequence semantics to boost the effectiveness of learning cross-modal joint embedding.  Third, our MSJE  approach can leverage multi-modal semantics in different stages of our semantic enhanced joint embedding learning. By incorporating  TFIDF semantics, sequence semantics and category semantics, MSJE can effectively regulate and optimize the cross-modal joint embedding learning by refining the assigned category labels and by utilizing the batch-hard triplet loss and modality alignment loss. 
Experiments show that our MSJE approach consistently outperforms the state of the art approaches for both image-to-recipe and recipe-to-image retrieval services on Recipe1M benchmark dataset with over 1 million food images and 800K cooking recipes.


%

\ifCLASSOPTIONcompsoc
  \section*{Acknowledgments}
\else
  \section*{Acknowledgment}
\fi

This work is partially supported by the USA National Science Foundation under Grants NSF 2038029, 1564097, and an IBM faculty award. The first author Zhongwei Xie has performed this work as a two-year visiting PhD student at Georgia Institute of Technology (2019-2021, under the support from China Scholarship Council (CSC) and Wuhan University of Technology.

\ifCLASSOPTIONcaptionsoff
  \newpage
\fi



\bibliographystyle{IEEEtran}
\bibliography{TSC}

%

\begin{IEEEbiography}[{\includegraphics[width=1in,height=1.25in,clip,keepaspectratio]{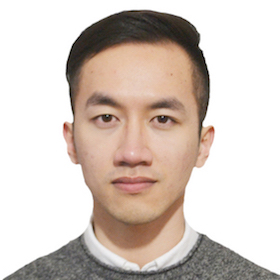}}]{Zhongwei Xie}
received his M.S degree in computer science from Wuhan University of Technology, Wuhan, China in 2017. He is currently a visiting student in Georgia Institute of Technology and also a Ph.D candidate in Wuhan University of Technology. His research interests include deep learning and cross-modal retrieval. 
\end{IEEEbiography}
\vspace{-30 pt} 
\begin{IEEEbiography}[{\includegraphics[width=1in,height=1.25in,clip,keepaspectratio]{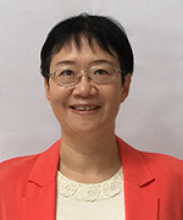}}]{Ling Liu}
is a Professor in the School of Computer Science at Georgia Institute of Technology. She directs the research programs in the Distributed Data Intensive Systems Lab (DiSL), examining various aspects of large scale big data-powered artificial intelligence (AI) systems, and machine learning (ML) algorithms and analytics, including performance, availability, privacy, security and trust. Prof. Liu is an elected IEEE Fellow, a recipient of IEEE Computer Society Technical Achievement Award (2012), and a recipient of the best paper award from numerous top venues, including IEEE ICDCS, WWW, ACM/IEEE CCGrid, IEEE Cloud, IEEE ICWS. Prof. Liu served on editorial board of over a dozen international journals, including the editor in chief of IEEE Transactions on Service Computing (2013-2016) and currently, the editor in chief of ACM Transactions on Internet Computing (TOIT). Prof. Liu has been given invited keynote speeches in many top tier venues in Big Data, AI and ML systems and applications, Cloud Computing, Services Computing, Privacy, Security and Trust. Her current research is primarily supported by USA National Science Foundation under CISE programs and IBM. 
\end{IEEEbiography}

\vspace{-30 pt} 
\begin{IEEEbiography}[{\includegraphics[width=1in,height=1.25in,clip,keepaspectratio]{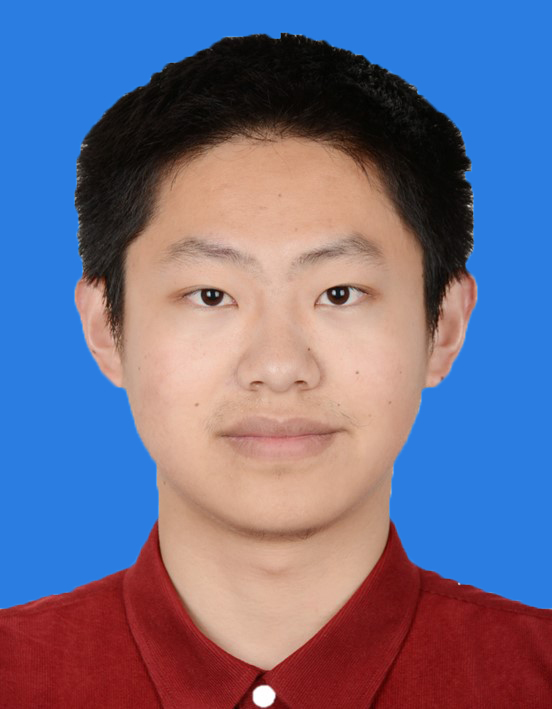}}]{Yanzhao Wu}
Yanzhao Wu is currently a PhD student in the School of Computer Science, Georgia Institute of Technology. He received his B.E. degree from the School of Computer Science and Technology, University of Science and Technology of China. His research interests are centered primarily on systems for machine learning and big data, and machine learning algorithms for optimizing systems.
\end{IEEEbiography}

\vspace{-30 pt} 
\begin{IEEEbiography}[{\includegraphics[width=1in,height=1.25in,clip,keepaspectratio]{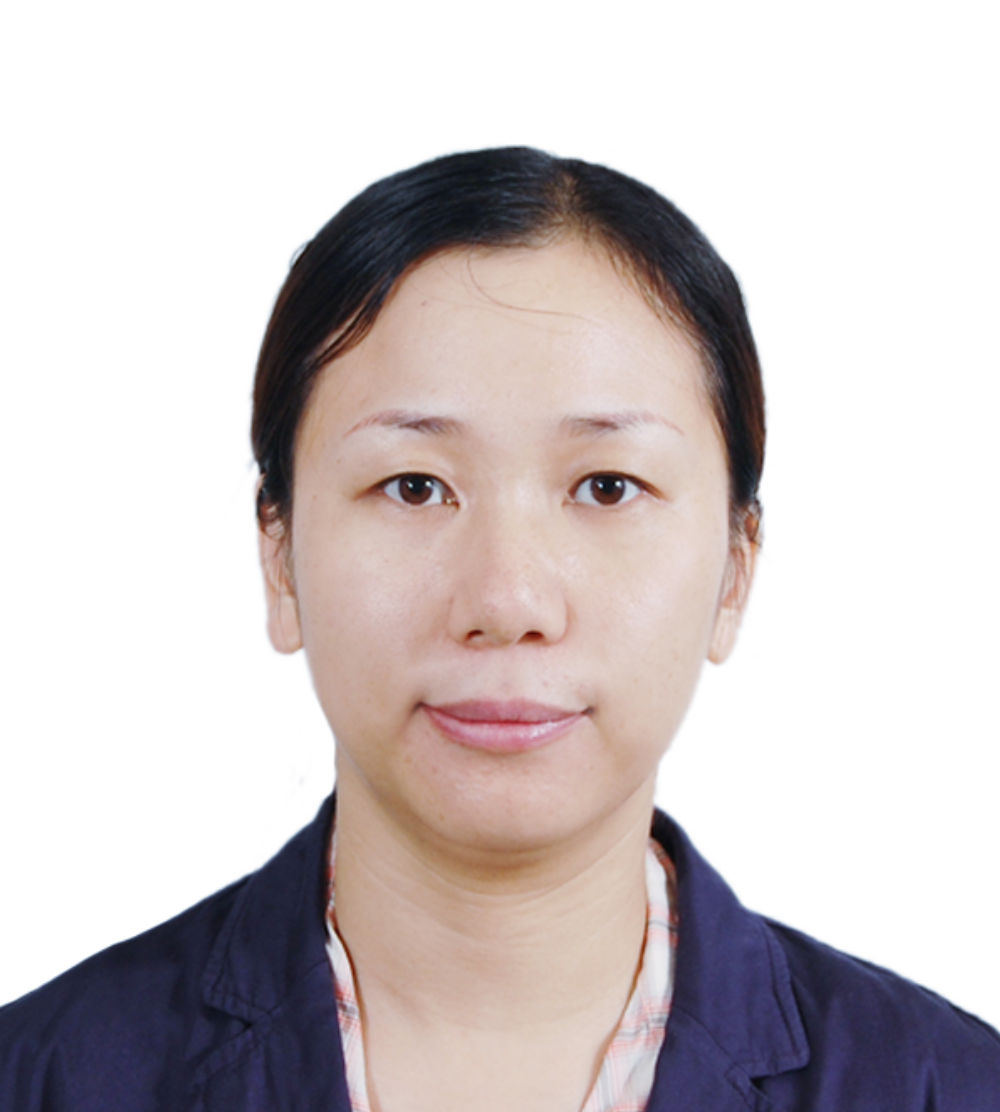}}]{Lin Li}
received her Ph.D in computer science from University of Tokyo, Japan in 2009. She is currently the Professor and PhD supervisor in Wuhan University of Technology. She visited University of Technology, Sydney as a visiting scholar between February 2014 and February 2015. Dr. Li's research interests mainly include but not limited to machine learning, text mining, information retrieval and recommender system. 
\end{IEEEbiography}
\vspace{-30pt} 
\begin{IEEEbiography}[{\includegraphics[width=1in,height=1.25in,clip,keepaspectratio]{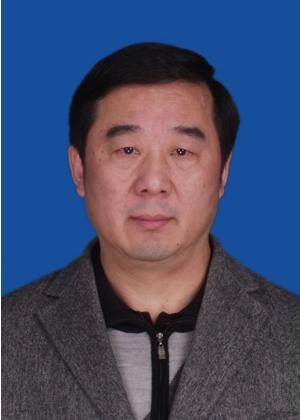}}]{Luo Zhong}
received his Ph.D in computer science from Wuhan University of Technology, China in 1996. He is currently the Professor and PhD supervisor in Wuhan University of Technology. Dr. Zhong's research interests mainly include but not limited to intelligent method and data mining.
\end{IEEEbiography}





\end{document}